\documentclass{article}

\usepackage[utf8]{inputenc} 
\usepackage[T1]{fontenc}    
\usepackage[pagebackref=false,breaklinks=true,colorlinks,bookmarks=false,citecolor=blue]{hyperref}

\usepackage{url}            
\usepackage{booktabs}       
\usepackage{amsfonts}       
\usepackage{nicefrac}       
\usepackage{microtype}      

\usepackage{amsmath}
\usepackage{amsthm}
\usepackage{amssymb}
\usepackage{bbm}
\usepackage{paralist}
\usepackage{enumitem}
\usepackage{xcolor}
\usepackage{mathtools}

\usepackage[ruled]{algorithm2e}

\usepackage{subcaption}

\newtheorem{theorem}{Theorem}

\newtheorem{lemma}{Lemma}

\newcommand{\EE}{\mathbb{E}}

\newcommand{\PP}{\mathbb{P}}

\newcommand{\indic}{\mathbbm{1} }

\newcommand{\bp}{\noindent{\bf Proof.}\ }
\newcommand{\ep}{\hfill $\Box$}

\newcommand{\als}[1]{ \begin{align*} #1  \end{align*}}
\newcommand{\eqs}[1]{ \begin{equation*} #1  \end{equation*}}

\newcommand{\Lp}{\left(}
\newcommand{\Rp}{\right)}

\renewcommand{\tilde}{\widetilde}

\renewcommand{\star}{*}

\newcommand{\kl}{\textnormal{kl}}

\title{An Optimal Algorithm\\ for Multiplayer Multi-Armed Bandits }

\author{Alexandre Proutiere and Po-An Wang\\
KTH, Royal Institute of Technology\\
Stockholm, Sweden}

\begin{document}

\maketitle

\begin{abstract}
The paper addresses the Multiplayer Multi-Armed Bandit (MMAB) problem, where $M$ decision makers or players collaborate to maximize their cumulative reward. When several players select the same arm, a collision occurs and no reward is collected on this arm. Players involved in a collision are informed about this collision. We present DPE (Decentralized Parsimonious Exploration), a decentralized algorithm that achieves the same regret as that obtained by an optimal centralized algorithm. Our algorithm has better regret guarantees than the state-of-the-art algorithm SIC-MMAB \cite{boursier2019}. As in SIC-MMAB, players communicate through collisions only. An additional important advantage of DPE is that it requires very little communication. Specifically, the expected number of rounds where players use collisions to communicate is finite. 
\end{abstract}

\section{The Multiplayer MAB problem}

In MMAB problems, there are $M$ independent decision makers. In each round, each decision maker selects an arm among the set ${\cal K}=\{1,\ldots,K\}$. $K$ is known to the decision makers, but they do not necessarily know $M$. In round $t$, when arm $k$ is selected, the potential collected reward is a random variable (independent of the rewards of the other arms) $X_k(t)$ with Bernoulli distribution with mean $\mu_k$. This reward is only collected by the decision maker if no other decision maker has selected $k$ in round $t$. Assume without loss of generality that $\mu_1 >\mu_2 >\ldots >\mu_K$, and that $K>M$ (if $K\le M$, the problem just boils down to making sure that each is played -- i.e., their expected rewards do not need to be learnt). We denote $\mu=(\mu_1,\ldots,\mu_K)$. When in round $t$, the decision maker $i$ selects $k$, she observes (1) whether her decision collides with those of other decision makers, and (2) $X_k(t)$ in the absence of collision. This feedback scenario is referred to as {\it collision sensing} in \cite{boursier2019}. 

A policy $\pi$ determines in each round which arm every decision maker will select. We are interested in distributed policies where each decision maker decides which arm to select independently. This choice depends on the available information to the decision maker: the past observed collisions and rewards. We denote by $k_i^\pi(t)$ the arm selected by the decision maker $i$ in round $t$.    

The optimal expected reward that can be collected in each round is $\sum_{k=1}^M\mu_k$ (when the $M$ best arms are played). Hence the expected regret up to round $T$ of a policy $\pi$ is defined as:
$$
R^\pi(T) =T \sum_{k=1}^M\mu_k - \sum_{t=1}^T\sum_{i=1}^M \mathbb{E}[\mu_{k_i^\pi(t)}].
$$ 

As in the classical bandit literature \cite{lai1985}, we say that a policy $\pi$ is {\it uniformly} good if it regret satisfies $R^\pi(T)=o(T^\alpha)$ for all $\alpha>0$ for any possible $\mu$. We know from \cite{venkat87} that any uniformly good policy $\pi$, centralized or not, satisfies:
\begin{equation}\label{eq:low}
\lim\inf_{T\to\infty} {R^\pi(T)\over \log(T)} \ge C(\mu):= \sum_{k>M} {\mu_M-\mu_k\over \kl(\mu_k,\mu_M)},
\end{equation}
where $\kl(a,b)$ denotes the KL divergence between two Bernoulli distributions of respective means $a$ and $b$. This result is a simple extension of the classical result derived by Lai and Robbins in \cite{lai1985}. \cite{venkat87} also presents a centralized policy achieving the above asymptotic regret lower bound. In this paper, we present a decentralized policy also achieving this fundamental regret limit.

\section{Decentralized Parsimonious Exploration}
 
We present DPE (Decentralized Parsimonious Exploration), a simple policy that achieves the asymptotic fundamental regret limit (\ref{eq:low}). The policy relies on the observation that in a MAB problem where the decision maker selects $M$ arms in each round (a model referred to as MAB with multiple plays \cite{venkat87}), an optimal algorithm consists in playing the $(M-1)$ best empirical arms and exploring using the remaining arm according to an optimal index policy, such as KL-UCB \cite{lai1987, garivier2011kl}. This observation that such {\it parsimonious} exploration suffices was already made and exploited in \cite{magureanu2015} for the design of learning-to-rank algorithms. It is powerful in the design of decentralized MMAB algorithm: Indeed, it implies that the exploration can be only performed by a single player, the so-called {\it leader}; the other players, referred to as the {\it followers}, just need to play the best empirical arms greedily. To this aim, the leader just needs to inform the followers when the set of the $M$ empirical arms changes -- and it can be done using collisions as proposed in \cite{boursier2019, boursier2019practical}. Note however that the communication protocol used in \cite{boursier2019} is complicated because players need to communicate their statistics of the arms. With the parsimonious exploration principle, the leader just needs to communicate the indexes of the best empirical arms.

\medskip

Next we present DPE in detail, and explain its advantages over the SIC-MMAB algorithm.

\subsection{Initialization phase} 

The first phase consists in coordinating the players. After this phase, a single player becomes the {\it leader}; this player is ranked first and is aware of this rank. The other players are {\it followers} and get to know their respective ranks $2,\ldots, M$. All players learn in passing the number of players $M$. After this phase they can  coordinate and avoid collisions except if they need collisions to communicate. SIC-MMAB also starts with such an initialization phase; this phase has by design a fixed duration $T_0=\lceil K\log(T)\rceil$, which implies in particular that its cost in terms of expected regret is $KM\log(T)$. In contrast, the initialization phase in DPE has a random duration: it lasts until all the objectives of the phase have been reached. The expected duration of DPE initialization phase is finite, and hence just generates a constant expected regret.

\medskip

DPE initialization phase consists of two sub-phases:

\medskip
\noindent
{\bf Orthogonalization.} This first sub-phase aims at assigning in a distributed manner $M$ different arms within $\{1,\ldots,K-1\}$ to the various players. In this sub-phase, the players maintain an internal state with values in $\{0,1\ldots,K-1\}$: when the state is '0', it means that the player is not satisfied, and still needs to find a free arm. When the state is '$k$', it means that the player manages to select arm $k$ without collision, and she will keep this state until the end of the sub-phase. The sub-phase consists in a sequence of blocks of $K+1$ rounds: in the first round of a block, players with state different than '0' select the arm corresponding to their state, and players with '0' state randomly select an arm in $\{0,1\ldots,K-1\}$. The $K$ remaining rounds of the block are used to communicate the outcomes of the first round. This communication is done by selecting arm $K$ and by observing collisions. More precisely, if a player is in state $k\neq 0$, then she selects arm $k$ except in the $k$-th round where she selects $K$. If a player is in state '0', she selects arm $K$ in the $K$ rounds. Note that as long as there is a player in state '0', collisions are experienced by all players in the $K$ last round of the block. Hence, all the players know that all players are satisfied when no collision is experienced in a block. When such a block occurs for the first time, the sub-phase terminates, and all players are aware of this termination. Further observe that the expected duration of this sub-phase is finite because it is obviously stochastically bounded by a geometric random variable (with mean that depends on $K$ and $M$ only).

\medskip
\noindent
{\bf Rank assignment.} After the orthogonalization sub-phase, all the players have different states in $\{1,\ldots, K-1\}$. The rank assignment sub-phase consists of $K-1$ blocks of $K-1$ rounds. In the $k$-th block, should a player be in state $k$, she sequentially selects arms $1,2,\ldots,K-1$; a player with state $j\neq k$ selects $j$ in the $K-1$ rounds. Note that for example, if no player has state 1, the first block will have no collision. When on the contrary, there is a player in state 1, all other players experience a single collision in the first block, and hence know that such a player exists. Thus after the $K-1$ blocks, all players get to know (i) the number $M$ of players, and (ii) the rank of their state (a player gets the rank 1 if no other player has a state smaller than hers).   

\medskip
The initialization phase has overall a finite expected duration, and we can hence ignore the regret it induces. Without loss of generality in the remaining of the paper, we assume that at the first round, all players have a known unique rank in $\{1,\ldots,M\}$. The rank-1 player is the leader and the other players are followers.  

\subsection{Exploration-exploitation phase}

In DPE, the leader is responsible for exploring and maintaining the set of the $M$ best empirical arms. Exploration is conducted using the following KL-UCB indexes. The index of arm $k$ in round $t$ is
$$
b_k(t) = \sup\{ q\ge 0: N_k(t)\kl(\hat{\mu}_k(t),q)\le f(t)\},
$$
where $f(t)=\log(t)+4\log\log(t)$, $N_k(t)$ denotes the number of times the leader has played arm $k$ up to round $t$, and $\hat{\mu}_k(t)$ is the empirical average of arm $k$ based on the rewards obtained before round $t$: for any $k$, $N_k(1)=0=\hat{\mu}_k(1)$, and for all $t>1$,
$$
N_k(t) = \sum_{s=1}^{t-1} \indic{\{\rho(s)=k\}},\ \ \ \hat{\mu}_k(t)={1\over N_k(t)}\sum_{s=1}^{t-1} \indic{\{\rho(s)=k\}}X_k(s),
$$
where $\rho(t)$ denotes the arm selected by the leader in round $t$. The leader is also responsible for communicating to the followers when the set ${\cal M}(t)$ of the $M$ best empirical arms changes. To this aim, she leverages collisions in the same manner as in SIC-MMAB. Each time ${\cal M}(t)$ changes, a communication phase is initiated by the leader, and this phase lasts a finite number of rounds. The algorithm is designed so that the expected number of times ${\cal M}(t)$ changes is finite, see Lemma \ref{lem3}. Hence we can ignore the communication cost, as it is sub-logarithmic. The followers just play different arms from ${\cal M}(t)$. Note that the followers do not need to communicate anything to the leader; in particular, the rewards they collect is not taken into account by the leader. Each communication phase has a fixed and finite duration, and is known to all players -- see Subsection \ref{subsec:comm} for detail. Hence without loss of generality, we ignore these periods of communication and we can assume that the leader communicates the new ${\cal M}(t)$ {\it instantaneously} whenever required. Communication phases do not impact the asymptotic regret of the algorithm.

The set of rounds is divided into blocks of $M$ rounds. In rounds belonging to the same block, the empirical means of the arms, the KL-UCB indexes, and the set of best empirical arms are kept constant. More precisely, the decisions made in one block are based on:
\begin{align*}
\hat{\nu}_k(t) &= \hat{\mu}_k(\lfloor {t\over M}\rfloor M),\quad d_k(t)= b_k(\lfloor {t\over M}\rfloor M),\quad k\in \{1,\ldots,K\},\\
{\cal N}(t)&= {\cal M}(\lfloor {t\over M}\rfloor M).
\end{align*}
At the beginning of a block, the leader updates the above variables. The block structure is designed so that (i) the leader gathers one sample of each of the $(M-1)$ best empirical arms, (ii) each follower selects each arm in ${\cal N}(t)$ once, and (iii) the leader explores only when the followers play the $(M-1)$ best empirical arms. Let us describe in more detail the DPE algorithm in more detail. 

\medskip
\noindent
{\bf Leader.} At the beginning of round $t$, if $t = 0(\hbox{mod }M)$, the leader updates $\hat{\nu}(t)$, $d(t)$, and  ${\cal N}(t)$. The set ${\cal N}(t)$ is ordered: ${\cal N}(t)=\{ \ell_{1}(t),\ldots,\ell_{M}(t)\}$. This order is arbitrary, but independent of the empirical means of the arms. In particular, the order is kept fixed even if the relative empirical means of the arms in ${\cal N}(t)$ evolve, so that the leader only needs to communicate to the followers when ${\cal N}(t)$ changes. Ordering ${\cal N}(t)$ is important to avoid collisions. In the following, we denote by $\hat{M}(t)$ the arm in ${\cal N}(t)$ with the smallest empirical mean.

If ${\cal N}(t)\neq {\cal N}(t-1)$, the leader communicates to the followers the identity of the arm leaving the set and that of the new arm that replaces it in ${\cal N}(t)$ (the rank of the new arm inherits that of the arm that left). 

The sequential arm selections made by the leader are as follows. In round $t$, define $m= [(t+1)(\hbox{mod }M)] +1$. If $\ell_m(t)\neq \hat{M}(t)$, then the leader selects $\rho(t)=\ell_{m}(t)$. If $\ell_m(t)= \hat{M}(t)$, then with probability 1/2, the leader selects arm $\hat{M}(t)$, and with probability 1/2, the leader plays an arm $k\notin {\cal N}(t)$ such that $d_k(t) > \hat{\nu}_{\hat{M}(t)}$, should such an arm exists, and plays $\hat{M}(t)$ otherwise.  

\medskip
\noindent
{\bf Followers.} The followers just exploit the knowledge of the leader: they play greedily different arms of ${\cal N}(t)$. More precisely, the follower with rank $i \in \{2,\ldots,M\}$ plays in round $t$ the arm $\ell_{{m}_i}(t)$ where $m_i =  [(t+i) \hbox{(mod } M)] +1$. 

The pseudo-code of the exploration-exploitation phase of the DPE algorithm is presented in Algorithm \ref{alg:main}.

\begin{algorithm}[htb]
	\SetAlgoLined
	\textbf{Initialization:} Set $\hat{\nu}(1)=d(1)=0$. Initialize the set of best empirical arms ${\cal N}(1)$  and $\hat{M}(1)$ arbitrarily.\\
	For round $t\ge 1$:\\
	{\bf Leader.}\\
	\qquad	1. If $t=0 (\hbox{mod }M)$, update $\hat{\nu}_k(t)$, $d_k(t)$ for each arm $k$, and $\hat{M}(t)$\\
	\qquad	\quad update the ordered set ${\cal N}(t)\gets\left\{\ell_{1}(t),\ell_{2}(t),\ldots,\ell_{M}(t)\right\}$\\
	\qquad  \quad (the set of the $M$ best empirical arms)\\
	\qquad    2. If ${\cal N}(t)\neq {\cal N}(t-1)$, communicate ${\cal N}(t)$ to the followers\\
	\qquad   3. ${\cal B}(t)\gets \left\{k\notin {\cal N}(t):d_k(t)\geq \hat{\nu}_{\hat{M}(t)}(t)\right\}$;\\
		\qquad  \quad  $m\gets \left[(t+1)(\hbox{mod } M)\right] +1$\\
	\qquad  \quad If (${\cal B}(t)=\emptyset$ or $\ell_m(t)\neq \hat{M}(t)$), $\rho(t)\gets \ell_{m}(t)$\\
		\qquad  \quad Else \\
		\qquad  \quad \quad w.p. $1/2$, $\rho(t)\gets  \hat{M}(t)$\\
		\qquad  \quad \quad w.p. $1/2$, $\rho(t)\gets  k $ where $k$ is drawn from ${\cal B}(t)$
		    uniformly\\
		\qquad  \quad  Select arm $\rho(t)$\\
		
{\bf Follower with rank $i\in \{2,\ldots,M\}$.}\\
		  \qquad $m_i\gets \left[(t+i)(\hbox{mod } M)\right] +1$, Select arm ${\ell}_{m_i}(t)$\\
    \caption{The DPE algorithm: Exploration-exploitation phase}~\label{alg:main}
\end{algorithm}

\subsection{Communication phases}\label{subsec:comm}

When ${\cal N}(t)\neq {\cal N}(t-1)$ has changed, the leader communicates the new ordered set ${\cal N}(t)$ as follows. She uses $M-1$ blocks of $M+K+1$ rounds. The $i$-th block is designed to communicate with the follower with rank $i+1$. For each block, the leader proceeds as follows. (i) In the first round, the leader selects the same arm as the follower to signal the beginning of the communication. (ii) the next $M$ rounds are used to communicate the rank $k$ in ${\cal N}(t)$ of the arm leaving ${\cal N}(t)$; this is done by only selecting the same arm as the follower in the $k$-th round. (iii) finally in a similar way, the leader uses the $K$ remaining rounds to communicate the index of the arm entering ${\cal N}(t)$. The new arm added to ${\cal N}(t)$ enters at the rank of the arm that leaves the set.  

Importantly, the followers continue to play according to the exploration-exploitation phase during the entire communication phase (until the leader has communicated to all followers). They change their selections only at the end of the communication phase. Note that since the followers know their rank, they know when the communication phase started and when it ends.

\section{Regret Analysis}

This section is devoted to the regret analysis of the DPE algorithm. We have:

\begin{theorem}\label{th1} For any $\mu$, the regret of $\pi=$DPE satisfies:
\[
\limsup_{T\rightarrow \infty}\frac{R^\pi(T)}{\log T}\leq \sum_{k>M}\frac{\mu_{M}-\mu_{k}}{\kl\left(\mu_k,\mu_M\right)}.
\]
\end{theorem}

To establish the result, we prove that the expected number of communication phases is finite. This is a consequence of Lemma \ref{lem3}. We also prove that the exploration-exploitation phase yields similar regret as the centralized KL-UCB algorithm, and hence minimizes the exploration of sub-optimal arms. The proof exploits the arguments used in \cite{magureanu2015} to establish a regret upper bound of a centralized algorithm for some MAB problems with multiple plays. 

\subsection{Preliminaries}

In the proof of Theorem \ref{th1}, we repeatedly use the following lemma. The latter is a simplified version of Lemma 5 in \cite{magureanu2015}). For completeness, we provide its proof in the appendix. In what follows, ${\cal F}_n$ denotes the $\sigma$-algebra generated by $(X_k(t),k\in [K], t\le n)$.

\begin{lemma}\label{lem:random}
Let $k\in [K]$, and $c > 0$. Let $H$ be a random set of rounds such that for all $n$, $\{n\in H\}\in {\cal F}_{n-1}$. Assume that there exists $(C_t)_{t\ge 0}$, a sequence of independent binary random variables, independent of all ${\cal F}_n, n\ge0$, such that for $n\in H$, $k$ is selected ($\rho(n)=k$) if $C_n=1$. Further assume that $\mathbb{P}[C_t=1]\ge c$, for any $t$. Then:
$$
\sum_{n\ge 1} \mathbb{P}[n\in H, | \hat\mu_k(n) - \mu_k | \ge \delta \} ]  \leq  2c^{-1} \left( 2c^{-1}  +  \delta^{-2} \right).
$$
\end{lemma}

In addition, we need some known results about the KL-UCB indexes. The following lemma is a direct consequence of Theorem 10 in \cite{garivier2011kl}.

%

\begin{lemma}~\label{cor:b<mu}
Under the DPE algorithm, we have:
$$
\sum_{n\geq 1}\mathbb{P}\left[d_k(n)<\mu_k\right]\leq C_0,
$$
where
$$
C_0\leq e M \sum_{s\geq 1}\lceil (\log (sM)+ 4\log(\log (sM) ))\log (sM)\rceil e^{-\log (sM)-4\log (\log (sM))}\leq 15.
$$
\end{lemma}

\subsection{Proof of Theorem \ref{th1}}

Let ${\cal M}^\star = \{1,\ldots,M\}$ be the set of the $M$ best arms. Further define $\delta_0= \min_{1\leq k\leq K-1} \frac{\mu_k-\mu_{k+1}}{2}$ as half of the minimum gap between the expected rewards of the arms. In what follows we choose $0<\delta <\delta_0$. We finally define for any $t\ge 1$, $m(t)=\left[(t+1)(\hbox{mod } M)\right] +1$.

\medskip
\noindent
We define the following sets of rounds:
\begin{itemize}
    \item[] $\mathcal{A}= \left\{n\geq 1: {\cal N}(n)\neq {\cal M}^\star\right\}$,
    \item[] $\mathcal{D}= \left\{ n\geq 1 :\exists  k \in {\cal N}(n) \text{ s.t. } \left|\hat{\nu}_k(n)-\mu_k\right|\geq \delta \right\}$,
    \item[] $ \mathcal{E}= \left\{n\geq 1:\exists k \in {\cal M}^*, d_k(n)< \mu_k\right\}$,
    \item[] $\mathcal{G}= \left\{n\geq 1: n\in \mathcal{A}\backslash (\mathcal{D}\cup \mathcal{E}),\exists k\in {\cal M}^*\backslash {\cal N}(n) \text{ s.t. } \left|\hat{\nu}_k(n)-\mu_k\right|\geq \delta\right\}$.
\end{itemize}

\medskip

\begin{lemma}\label{lem3}
$(\mathcal{A}\cup\mathcal{D})\subseteq (\mathcal{D}\cup \mathcal{E}\cup \mathcal{G}).$ As a consequence, we have 
\[
\mathbb{E}\left[\left|\mathcal{A}\cup \mathcal{D}\right|\right]\leq \mathbb{E}\left[\left|\mathcal{D}\right|\right]+\mathbb{E}\left[\left|\mathcal{E}\right|\right]+\mathbb{E}\left[\left|\mathcal{G}\right|\right].
\]
\end{lemma}

\medskip
\bp Let $n\in \mathcal{A}\backslash (\mathcal{D}\cup \mathcal{E})$. We show that $n\in\mathcal{G}$. Since $n\notin \mathcal{D}$, $\forall k\in {\cal N}(n)$, we have
\begin{equation}~\label{eq:n notin d}
    \left|\hat{\nu}_k(n)-\mu_k\right| <\delta. 
\end{equation}
Moreover, $n\in \mathcal{A}$. Hence there exists  
$j\in {\cal M}^*\backslash {\cal N}(n)$ such that 
\begin{equation}~\label{eq:wrong j}
    \hat{\nu}_j(n)<\hat{\nu}_k(n)\text{ for some }k\in {\cal N}(n)\backslash {\cal M}^*. 
\end{equation}
Combining (\ref{eq:n notin d}) and (\ref{eq:wrong j}) leads to
$
\hat{\nu}_j(n)<\hat{\nu}_k(n)\leq \mu_k+\delta\leq \mu_M-\delta\leq \mu_j-\delta.
$
The last two inequalities are due to our assumption that $j\geq M>k$ and $\delta<\delta_0$. 
It implies $\left|\hat{\nu}_j(n)-\mu_j\right|\geq \delta$ and thus, since $n \notin\mathcal{D}\cup \mathcal{E}$, $n\in \mathcal{G}$. 
Therefore, $\mathcal{A}\cup\mathcal{D}\subseteq \mathcal{D}\cup \mathcal{E}\cup \mathcal{G}$.
\ep

\medskip

\begin{lemma}\label{lem:DEG}
We have: $\mathbb{E}\left[\left|\mathcal{D}\right|\right]+\mathbb{E}\left[\left|\mathcal{E}\right|\right]+\mathbb{E}\left[\left|\mathcal{G}\right|\right]\leq 8MK^2(6K+\delta^{-2}).$
\end{lemma}

\medskip
\bp
We upper bound each term. \\
(a)  We show that $\mathbb{E}\left[\left|\mathcal{D}\right|\right]<4MK(4+\delta^{-2})$. For each $k\in \left\{1,2,\ldots,K\right\} $, define $\mathcal{D}_k=\left\{n\geq 1:k\in {\cal N}(n),\left|\hat{\nu}_k(n)-\mu_k\right|\geq \delta\right\}$. 
We have $\mathcal{D}=\cup_{1\leq k\leq K}\mathcal{D}_k$. 
Let us fix $k\in \left\{1,2,\ldots,K\right\}$ and split $\mathcal{D}_k$ into two sets,
\begin{align*}
    \mathcal{D}_{k,1}=\left\{n\in \mathcal{D}_k: \ell_{m(n)}(n)=k \right\},\\
    \mathcal{D}_{k,2}=\left\{n\in \mathcal{D}_k: \ell_{m(n)}(n)\neq k\right\}.
\end{align*}
We first upper bound the expected cardinality of $\mathcal{D}_{k,1}$. To this aim, notice that if $n\in \mathcal{D}_{k,1}$, then $|\hat{\mu}_k(n)-\mu_k|\ge \delta$. Indeed, by design of the algorithm, $k$ is not played between round $\lfloor {n\over M}\rfloor M$ and round $n-1$, hence $\hat{\mu}_k(n)=\nu_k(n)$. We deduce that:
$$
{\cal D}_{k,1}=\{n\ge 1: |\hat{\mu}_k(n)-\mu_k|\ge \delta, \ell_{m(n)}(n)=k\}.
$$
We can now apply Lemma~\ref{lem:random} with $H=\{ n\ge 1: \ell_{m(n)}(n)=k\}$. Note that $\{ n\in H\}\in {\cal F}_{n-1}$. By design of the algorithm, for $n\in H$, $k$ is selected with probability at least $c=1/2$. Thus: $\mathbb{E}\left[\left|\mathcal{D}_{k,1}\right|\right]\leq 4(4+\delta^{-2})$. 

\medskip
\noindent
To upper bound the expected cardinality of $\mathcal{D}_{k,2}$, observe that since the DPE algorithm operates by blocks of $M$ rounds (i.e., $\hat{\mu}(t)$, $b(t)$, and ${\cal M}(t)$ do not change over $M$ consecutive rounds), when $n\in \mathcal{D}_{k,2}$, there exists a round $p$ such that $|n-p|<M$ ($p$ belongs to the same block as $n$) and such that $p\in \mathcal{D}_{k,1}$. Hence $|\mathcal{D}_{k,2}|\le (M-1) |\mathcal{D}_{k,1}|$. 

We have established that $\mathbb{E}\left[\left|\mathcal{D}_k\right|\right]\le 4M(4+\delta^{-2})$, and thus $\mathbb{E}\left[\left|\mathcal{D}\right|\right]<4MK(4+\delta^{-2})$.

\medskip
\noindent
(b) We show that $\mathbb{E}\left[\left|\mathcal{E}\right|\right]<15M$. We apply Lemma~\ref{cor:b<mu} for each arm $k\in\left\{1,2,\ldots,M\right\}$, so $\sum_{n\geq 0}\mathbb{P}\left[d_k(n)<\mu_k\right]\leq 15$.  We simply deduce that:
$$
\mathbb{E}\left[ \left|\mathcal{E}\right|\right]\leq 15 M.
$$

\noindent
(c) We show that $\mathbb{E}\left[\left|\mathcal{G}\right|\right]<4K^2M(4K+\delta^{-2})$.\\
Define $\mathcal{G}_k=\left\{n\geq 1:n\in \mathcal{A}\backslash(\mathcal{D}\cup\mathcal{E}),k\notin {\cal N}(n),\left|\hat{\nu}_k(n)-\mu_k\right|\geq \delta\right\}$ for all $k\in {\cal M}^*$. Then $\mathcal{G}\subseteq \bigcup_{k\leq M} \mathcal{G}_k$. \\
Fix $k\in {\cal M}^\star$, and let $n\in \mathcal{G}_k$. Since $n\notin \mathcal{D}$, we have for all $j\in {\cal N}(n)$, $|\hat{\nu}_{j}(n)-\mu_{j}|< \delta$. Now let $j^\star =\max\{j: j\in {\cal N}(n)\}$. We have $j^\star>M$ (since $n\in {\cal A}$), which implies that:

\begin{equation}~\label{eq:border M}
    \hat{\nu}_{j^\star}(n)< \mu_{j^\star}+\delta \leq \mu_{M+1}+\delta< \frac{\mu_{M+1}+\mu_M}{2}.
\end{equation}
The last inequality follows from the definition of $\delta< \frac{\mu_{M+1}-\mu_M}{2}$. Furthermore, since $n\notin\mathcal{E}$, 
\begin{equation}~\label{eq:b biger than mean}
    d_k(n)\geq \mu_k.
\end{equation}
Combining (\ref{eq:border M}) and (\ref{eq:b biger than mean}), we get 
$$
d_k(n)\geq \mu_k\geq (\mu_{M+1}+\mu_M)/2\geq \hat{\nu}_{j^\star}(n)\geq \hat{\nu}_{\hat{M}(n)},
$$
where the last inequality stems from the fact that $j^\star\in {\cal N}(n)$. Observe then that such a round $n$, by the design of algorithm, arm $k$ will be selected with probability at least $1/2K$ when $\ell_{m(n)}(n)=\hat{M}(n)$ (exploration rounds). Next we split $\mathcal{G}_k$ into the following two sets: 
\begin{align*}
    \mathcal{G}_{k,1}=\left\{n\in \mathcal{G}_k: \ell_{m(n)}(n)=\hat{M}(n) \right\},\\
    \mathcal{G}_{k,2}=\left\{n\in \mathcal{G}_k: \ell_{m(n)}(n)\neq \hat{M}(n)\right\}.
\end{align*}
For the set $\mathcal{G}_{k,1}$, we apply Lemma~\ref{lem:random} with $H= \mathcal{G}_{k,1}$, $c= 1/2K$, and deduce that $\mathbb{E}\left[\left|\mathcal{G}_{k,1}\right|\right]\leq 4K(4K+\delta^{-2})$. 

\medskip
\noindent
For the set $\mathcal{G}_{k,2}$, again since the algorithm works in blocks, when $n\in \mathcal{G}_{k,2}$, there exists a round $p$ such that $|n-p|<M$ ($p$ belongs to same block as $n$) and such that $p\in \mathcal{G}_{k,1}$. Hence, $\mathbb{E}\left[\left|\mathcal{G}_{k,2}\right|\right]\leq (M-1)\mathbb{E}\left[\left|\mathcal{G}_{k,1}\right|\right]$, and  
\[
\mathbb{E}\left[\left|\mathcal{G}\right|\right]\leq \sum_{k=1}^K \left(\mathbb{E}\left[\left|\mathcal{G}_{k,1}\right|\right]+\mathbb{E}\left[\left|\mathcal{G}_{k,2}\right|\right]\right)\leq 4K^2M(4K+\delta^{-2}).
\]
\ep

\begin{lemma}~\label{lem:C}
Given $T\geq 1$ and some $k\in\left\{M+1,\ldots,K\right\}$, we define $\mathcal{C}_k=\left\{n\leq T,n\notin \mathcal{A}\cup \mathcal{D},\rho(n)=k\right\}$. We show that $$\mathbb{E}\left[\left|\mathcal{C}_k\right|\right]\leq  \frac{\log T+4\log(\log T)}{\kl\left(\mu_k+\delta,\mu_M-\delta\right)}+4+2\delta^{-2}.$$

\end{lemma}

\bp Define the counter $c(n)= \sum_{t=1}^n \mathbbm{1}_{\left\{t\in \mathcal{C}_k\right\}}$, which is the number of rounds in $\mathcal{C}_k$ before round $n$ and $t_0 = \left(\log T+4\log(\log T)\right)/\kl\left(\mu_k+\delta,\mu_M-\delta\right)$. \\
Define two subsets of $\mathcal{C}_k$ as 
\begin{align*}
\mathcal{C}_{k,1} & = \left\{n\in\mathcal{C}_k:\left|\hat{\nu}_k(n)-\mu_k\right|\geq \delta\right\},\\
\mathcal{C}_{k,2} & = \left\{ n\in\mathcal{C}_k:c(n)< t_0 \right\}.
\end{align*}

We first show that $\mathcal{C}_k\subseteq \mathcal{C}_{k,1}\cup \mathcal{C}_{k,2}$. Let $n\in \mathcal{C}_k\backslash (\mathcal{C}_{k,1}\cup \mathcal{C}_{k,2})$. Since $n\notin \mathcal{C}_{k,2}$, 
\begin{equation}\label{eq:N>t_0}
    N_k(n)\geq c(n)\geq t_0.
\end{equation}
Then $n\notin \mathcal{A}$ implies that ${\cal M}^*={\cal N}(n)$. 
Hence $\rho(n)=k$ can only happen when 
\begin{equation}\label{eq:border}
    d_k(n)\geq \hat{\nu}_{\hat{M}(n)}(n)=\hat{\nu}_{M}(n).
\end{equation} 
Moreover, $n\notin \mathcal{D}$ implies that 
\begin{equation}~\label{eq:well M}
    \hat{\nu}_M(n)> \mu_M-\delta.
\end{equation}
Finally $n\notin \mathcal{C}_{k,1}$ and $\delta < \min_{1\leq k\leq K-1} \frac{\mu_k-\mu_{k+1}}{2}$ imply that 
\begin{equation}~\label{eq:well k}
    \hat{\nu}_k(n) < \mu_k +\delta < \mu_M-\delta.
\end{equation}
Combining the above arguments, we get:
\begin{eqnarray}~\label{eq:contradic}
t_0\kl\left(\hat{\nu}_k(n),\mu_M-\delta\right)&\leq &N_{k}(n)\kl\left(\hat{\nu}_k(n),\mu_M-\delta\right)\\\nonumber
&\leq & N_{k}(n)\kl\left(\hat{\nu}_k(n), d_k(n)\right)\\\nonumber
&\leq &\log T+4\log(\log T).\nonumber
\end{eqnarray}
The first inequality follows from~(\ref{eq:N>t_0}); the second inequality stems from~(\ref{eq:border})-(\ref{eq:well M})-(\ref{eq:well k}) and the fact that $y\mapsto\kl\left(x, y\right)$ is an increasing function when $0<x<y<1$; the last inequality is obtained by definition of $d_ k(n)$, and by the fact that $N_k(n)=N_k(\lfloor {n\over M}\rfloor M)$ since an arm $k$ can only be selected once per block. Replacing $t_0$ by its value in the above inequality, we finally obtain:
$$
\kl\left(\hat{\nu}_k(n),\mu_M-\delta\right)\leq \kl\left(\mu_k+\delta,\mu_M-\delta\right).
$$
Now observe that $x\mapsto\kl\left(x, y\right)$ is a decreasing function when $0<x<y<1$. We conclude that $\hat{\nu}_k(n)\geq \mu_k+\delta$ which contradicts the assumption that $n\notin \mathcal{C}_{k,1}$. Hence, $\mathcal{C}_k=\mathcal{C}_{k,1}\cup\mathcal{C}_{k,2}$. 

\medskip
\noindent
To complete the proof of the lemma, we upper bound $\mathbb{E}\left[\left|\mathcal{C}_{k,1}\right|\right]$ and $\mathbb{E}\left[\left|\mathcal{C}_{k,2}\right|\right]$.\\
For $\mathbb{E}\left[\left|\mathcal{C}_{k,1}\right|\right]$: we apply Lemma~\ref{lem:random} with $c= 1$, $H= \mathcal{C}_{k,1}$, and get $\mathbb{E}\left[\left|\mathcal{C}_{k,1}\right|\right]\leq 4+2\delta^{-2}$. 

\medskip
\noindent
For $\mathbb{E}\left[\left|\mathcal{C}_{k,2}\right|\right]$: if $n\in \mathcal{C}_{k,2}$, $c(n)\leq t_0$ and $c(n)$ is incremented by +1. 
Therefore, 
$$
\mathbb{E}\left[\left|\mathcal{C}_{k,2}\right|\right]\leq t_0= \frac{\log T+4\log (\log T)}{\kl\left(\mu_k+\delta, \mu_M-\delta\right)}.
$$

\medskip
\noindent
We have proved that: 
\[
\mathbb{E}\left[\left|\mathcal{C}_{k}\right|\right]\leq \frac{\log T+4\log (\log T)}{\kl\left(\mu_k+\delta, \mu_M-\delta\right)}+4+2\delta^{-2}.
\]
\ep

\medskip
\noindent
{\bf Proof of Theorem~\ref{th1}.} As already mentioned, the initialization phase generates a finite expected regret, and is hence ignored here. The expected regret can be bounded as follows:
\begin{align*}
R^\pi (T) &\leq 4KM\mathbb{E}\left[\left|\mathcal{A}\right|\right]+M\mathbb{E}\left[\left|\mathcal{A}\cup \mathcal{D}\right|\right]+\sum_{k>M}\left(\mu_M-\mu_k\right)\mathbb{E}\left[\left|\mathcal{C}_k\right|\right]
\end{align*}
The first term corresponds to an upper bound of the regret induced by communication rounds. Indeed, the number of rounds per communication phase is $(M-1)(M+K+1)\le 2KM$. In addition, note that the number of communication phases can be bounded as 
$$
\left|\left\{{t\ge 2:{\cal N}(t)\neq {\cal N}(t-1)}\right\}\right|\le  2 \left|\mathcal{A}\right|.
$$
Applying Lemmas~\ref{lem3} and \ref{lem:DEG}, we get: 
$$
4KM\mathbb{E}\left[\left|\mathcal{A}\right|\right]+M\mathbb{E}\left[\left|\mathcal{A}\cup \mathcal{D}\right|\right]\le 8K^2M^2(4K+1)(6K+\delta^{-2}).
$$
Hence, Lemma \ref{lem:C} yields:
Thus:
\[
\limsup_{T\rightarrow \infty}\frac{R^\pi(T)}{\log T}\leq \sum_{k>M}^K\frac{\mu_{M}-\mu_{k}}{\kl\left(\mu_k+\delta,\mu_M-\delta\right)}.
\]
The theorem is obtained by letting $\delta$ tend to 0.
\ep

\bibliographystyle{plain}
\bibliography{ref,vr2017}

\appendix

\section{Concentration lemmas}

\begin{lemma}\label{lem:concentr}
Let $({\cal F}_n)_{n\ge 0}$ a sequence of increasing $\sigma$-algebras, and denote by ${\cal G}=({\cal F}_{n-1},n\ge 1)$ a corresponding filtration. Let $(X_n)_{n\ge 0}$ be a sequence of independent random variables. Assume that with for any $n$, $X_n\in [0,1]$ is ${\cal F}_n$-measurable. Let $n_0\ge 1$ and $T\ge n_0$ two integers. We define the {\it partial} empirical sum $S_n = \sum_{t=n_0}^{n-1} B_t (X_t - \mathbb{E}[X_n])$, where for any $t\ge 1$, $B_t \in \{0,1\}$ is ${\cal F}_{t-1}$-measurable. Further define $t_n = \sum_{t=n_0}^{n-1} B_t$. Define $\phi \in \{n_0,\dots,T+1\}$ a ${\cal G}$-stopping time (i.e., $\{ \phi = t\} \in {\cal F}_{t-1}$) such that either $t_{\phi} \geq \zeta s$ or $\phi = T+1$, for some $\zeta >0$. 
 Then we have, for all $\delta>0$:
 $$
 \PP[ S_{\phi}  \geq t_{\phi} \delta \;,\;  \phi \leq T  ] \leq \exp( -2 \zeta s \delta^2).
 $$
 As a consequence:
 $$
 \PP[ | S_{\phi} | \geq t_{\phi} \delta \;,\;  \phi \leq T  ] \leq 2 \exp( -2 \zeta s \delta^2).
 $$
\end{lemma}

\bp Let $\delta, \lambda > 0$, and define $G_n = \exp( \lambda(S_n - \delta t_n)  ) \indic \{n \leq T \}$. We have that:
	\als{
	\PP[S_{\phi}  \geq t_{\phi} \delta \;,\;  \phi \leq T  ] = \PP[ \exp( \lambda(S_{\phi}  - \delta t_{\phi} ) ) \indic \{\phi \leq T \} \geq 1] = \PP[ G_{\phi}  \geq 1] \leq \EE[ G_{\phi} ].
	}
Next we provide an upper bound for $\mathbb{E}[G_{\phi}]$. We define the following quantities:
\begin{align*}
Y_t &=  B_t [  \lambda (X_t - \mathbb{E}[X_t])  - \lambda^2/8  ] \\
\tilde{G}_n &=  \exp \Lp \sum_{t=n_0}^n Y_t \Rp \indic \{n \leq T \}.
\end{align*}
We have $G_n = \tilde{G}_n \exp (- t_n ( \lambda \delta - \lambda^2 /8 ) )$, and setting $\lambda = 4 \delta$: $G_n  = \tilde{G}_n \exp(-2 t_n \delta^2 )$.
Using the fact that $t_{\phi} \geq \zeta s$ if $\phi \leq T$,  we can upper bound $G_{\phi}$ by:
	\eqs{
	G_{\phi} =  \tilde{G}_{\phi} \exp(-2 t_{\phi} \delta^2) \leq  \tilde{G}_{\phi}  \exp(-2 \zeta s \delta^2).
	}
Note that the above inequality holds even when $\phi = T+1$, since $G_{T+1} = \tilde{G}_{T+1} = 0$. Hence:
		\eqs{
	\EE[ G_{\phi} ] \leq \EE[ \tilde{G}_{\phi} ] \exp(-2 \zeta s \delta^2).
	}
We prove that $(\tilde{G}_n)_n$ is a ${\cal G}$-super-martingale. We have that $\EE[ \tilde{G}_{T+1} | {\cal F}_{T-1}] = 0 \leq \tilde{G}_{T}$. For $n \leq T-1$, since $B_{n}$ is ${\cal F}_{n-1}$ measurable:
	\eqs{
	\EE[ \tilde{G}_{n+1} | {\cal F}_{n-1}] = \tilde{G}_n (  (1 - B_{n}) +   B_{n} \EE[ \exp(\lambda(X_n-\mathbb{E}[X_n])- \lambda^2 /8) ] )  . 
	}
As in \cite{Hoeffding1963}[eq. 4.16], since $X_n \in [0, 1]$, we have:
	\eqs{
	\EE[ \exp(\lambda (X_n - \mathbb{E}[X_n]))  ] \leq \exp(\lambda^2/8 ),
	}
and hence $(\tilde{G}_n)_n$ is indeed a ${\cal G}$-supermartingale: $\EE[ \tilde{G}_{n+1} | {\cal F}_{n-1}] \leq \tilde{G}_n$. Since $\phi \leq T+1$ almost surely, and $(\tilde{G}_n)_n$ is a supermartingale, Doob's optional stopping theorem yields: $\EE[ \tilde{G}_{\phi}] \leq \EE[\tilde{G}_{n_0-1}] = 1$, and so
	\als{
		 \PP[& S_{\phi}  \geq t_{\phi} \delta  ,  \phi \leq T  ] \leq \mathbb{E}[G_\phi] \le \EE[ \tilde{G}_{\phi} ] \exp(-2 \zeta s \delta^2) \leq \exp(-2 \epsilon s \delta^2).
	}
	which concludes the proof. The second inequality is obtained by symmetry.

\ep

\medskip

\begin{lemma}\label{lem:concentr2}
Let $({\cal F}_n)_{n\ge 0}$ a sequence of increasing $\sigma$-algebras, and denote by ${\cal G}=({\cal F}_{n-1},n\ge 1)$ a corresponding filtration. 
Let $(X_n)_{n\ge 0}$ be a sequence of independent random variables. Assume that with for any $n$, $X_n\in [0,1]$ is ${\cal F}_n$-measurable. Let $n_0\ge 1$ and $T\ge n_0$ two integers. We define the {\it partial} empirical sum $S_n = \sum_{t=n_0}^{n-1} B_t (X_t - \mathbb{E}[X_t])$, where for any $t\ge 1$, $B_t \in \{0,1\}$ is ${\cal F}_{t-1}$-measurable. We assume that for all $t\ge 1$, almost surely, $B_t \ge \bar{B}_tC_t$, where $\bar{B}_t\in \{0,1\}$ are ${\cal F}_{t-1}$-measurable, and $(C_t)_{t\ge 0}$ are independent, independent of all ${\cal F}_n, n\ge0$, and such that  $\mathbb{P}[C_t=1]\ge c>0$.

Further define $t_n = \sum_{t=n_0}^{n-1} B_t$ and $c_n = \sum_{t=n_0}^{n-1} \bar{B}_t$. Define $\phi \in \{n_0,\dots,T+1\}$ a ${\cal G}$-stopping time (i.e., $\{ \phi = t\} \in {\cal F}_{t-1}$) such that either $c_{\phi} \geq s$ or $\phi = T+1$. 
 Then for all $\epsilon>0$ and $\delta>0$, we have:
 $$
 \PP[ S_{\phi}  \geq t_{\phi} \delta \;,\;  \phi \leq T  ] \leq e^{-2s\epsilon^2c^2}+e^{-2c(1-\epsilon)s\delta^2}.
 $$
 As a consequence:
 $$
 \PP[ | S_{\phi} | \geq t_{\phi} \delta \;,\;  \phi \leq T  ] \leq 2(e^{-2s\epsilon^2c^2}+e^{-2c(1-\epsilon)s\delta^2}).
 $$
\end{lemma}

\medskip
\bp Let $\epsilon, \delta >0$. Assume that $S_\phi \ge \delta t_\phi$. Then:
\begin{align*}
\hbox{either } & (a) \ \ \ \ t_\phi \le c(1-\epsilon)c_\phi,\\
\hbox{or }& (b) \ \ \ \ S_\phi\ge \delta t_\phi\hbox{ and }t_\phi > c(1-\epsilon)c_\phi.
\end{align*}

In case (a): if $\phi\le T$, 
$$
\sum_{t=n_0}^{\phi} \bar{B}_tC_t \le \sum_{t=n_0}^{\phi} {B}_t=t_\phi \le c(1-\epsilon)c_\phi=c(1-\epsilon)\sum_{t=n_0}^{\phi} \bar{B}_t.
$$
We deduce that: 
$$
\sum_{t=n_0}^{\phi} \bar{B}_t(C_t-c)\le -c\epsilon  \sum_{t=n_0}^{\phi} \bar{B}_t.
$$
Thus, since $\mathbb{E}[C_t]\ge c$, applying Hoeffding's inequality,
\begin{align*}
\mathbb{P}[t_\phi \le c(1-\epsilon)c_\phi, \phi\le T] & \le \mathbb{P}\left[ \sum_{t=n_0}^{\phi} \bar{B}_t(C_t-\mathbb{E}[C_t])\le -c\epsilon  \sum_{t=n_0}^{\phi} \bar{B}_t, \phi\le T\right]\\
&\le e^{-2s\epsilon^2c^2}.
\end{align*}

Now to upper bound the probability of (b) to occur, we define the following ${\cal G}$-stopping time:
$$
\phi' = \left\{
\begin{array}{ll}
\phi, & \hbox{ if }\ t_\phi > c(1-\epsilon)c_\phi,\\
T+1, & \hbox{ otherwise.}
\end{array}
\right.
$$
$\phi'$ is indeed a ${\cal G}$-stopping time, because for any $n\le T$, 
\begin{align*}
\{ \phi' = n\} & = \{ \phi=n, t_\phi > c(1-\epsilon)c_\phi \}\\
& = \{ \phi=n, t_n > c(1-\epsilon)c_n \} \in {\cal F}_{n-1},
\end{align*}
since $t_n$ and $c_n$ are ${\cal F}_{n-1}$-measurable. We apply Lemma \ref{lem:concentr} with $\phi'$ and $\zeta s \leftarrow c(1-\epsilon)s$ (indeed, if (b) and $\phi\le T$ hold, $t_{\phi'} > c(1-\epsilon)c_{\phi} \ge  c(1-\epsilon)s$), and conclude that:
\begin{align*}
\mathbb{P}[S_{\phi}\ge t_\phi\delta, t_\phi> c(1-\epsilon)c_\phi, \phi\le T]& \le \mathbb{P}[S_{\phi'}\ge t_\phi\delta, \phi'\le T]\\
& \le e^{-2c(1-\epsilon)s\delta^2}.
\end{align*}
Combining the analysis of cases (a) and (b), we get:
$$
\mathbb{P}[S_{\phi}\ge t_\phi\delta,\phi\le T] \le e^{-2s\epsilon^2c^2} +  e^{-2c(1-\epsilon)s\delta^2}.
$$
The second inequality is obtained by symmetry. \ep

\medskip
\noindent
{\bf Proof of Lemma \ref{lem:random}.} Let $s\ge 1$, define the ${\cal G}$-stopping time $\phi_s$ such that $\sum_{t=0}^{\phi_s}1_{\{n\in H\} }=s$. $\phi_s$ is the round corresponding to the $s$-th times the round belongs to $H$. Denote by $B_t=\bar{B}_tC_t$ where $\bar{B}_t=1_{\{t\in H\} }$. We can apply Lemma \ref{lem:concentr2}, and get:
$$
 \PP[ | \hat\mu_k(\phi_s) - \mu_k |\ge \delta \;,\;  \phi_s \leq T  ] \leq 2(e^{-2s\epsilon^2c^2}+e^{-2c(1-\epsilon)s\delta^2}).
$$
Now observe that: $\{ n\in H, | \hat\mu_k(n) - \mu_k | \ge \delta\} =\cup_s\{  | \hat\mu_k(\phi_s) - \mu_k | \ge \delta ;,\;  \phi_s \leq T\}$. A union bound yields:
\begin{align*}
\sum_{n\ge 1} \mathbb{P}[n\in H, | \hat\mu_k(n) - \mu_k | \ge \delta \} ] & \le \sum_{s\ge 1} 2(e^{-2s\epsilon^2c^2}+e^{-2c(1-\epsilon)s\delta^2})\\
&\le c^{-1} \left( {1\over \epsilon^2 c} + {1\over \delta^2(1-\epsilon)} \right).
\end{align*}
To establish the last inequality, we have used $\sum_{s\ge 1}e^{-ws}\le 1/w$ when $w>0$.\ep

\end{document}